%% file: ad-main.tex
\documentclass[sigconf,screen,nonacm]{acmart}

\usepackage{amsmath}
\usepackage{array}
\usepackage{booktabs}
\usepackage{enumitem}
\usepackage{tabularx}
\usepackage{colortbl}
\usepackage{pifont}
\usepackage{algorithm}
\usepackage{algpseudocode}

\newcommand{\cmark}{\ding{51}}
\newcommand{\xmark}{\ding{55}}

\setcopyright{none}
\renewcommand\footnotetextcopyrightpermission[1]{}
\title[AndroidDaily]{AndroidDaily: A Verifiable Benchmark for Mobile GUI Agents on Real-World Closed-Source Applications}

\author{
  YiFan Sui$^{1,2,*}$, Xin Huang$^{2,3,*}$, Hongbing Li$^{1,2}$, Fang Xu$^2$, Jiahe Lv$^2$,
  Haolong Yan$^{1,2}$, Yeqing Shen$^2$, Litao Liu$^2$, Zhimin Fan$^2$, Ziyang Meng$^2$, Jia Wang$^2$, Junbo Qi$^2$,
  Kaijun Tan$^2$, Zheng Ge$^2$, Xiangyu Zhang$^2$, Daxin Jiang$^2$, Osamu Yoshie$^3$
}
\thanks{*Both authors contributed equally to this research.}
\thanks{Preprint} 

\affiliation{%
  \institution{
    \vspace{0.3cm}
    $^1$Beijing University of Posts and Telecommunications \quad
    $^2$StepFun \quad
    $^3$Waseda University
  }
  \country{}
}

\acmConference[MM '26]{ACM Multimedia 2026}{November 2026}{Rio de Janeiro, Brazil}

\begin{document}

\begin{abstract}
\input{body/0-abstract}
\end{abstract}



\maketitle

\input{body/1-intro}

\input{body/2-related-work}
\input{body/3-bench0329}
\input{body/4-experimental-setup}
\input{body/5-limitations}
\input{body/5-conclusion}

\bibliographystyle{ACM-Reference-Format}
\bibliography{sample-base}


\end{document}

%% file: body/0-abstract.tex
The rapid development of GUI foundation models and mobile GUI agents has spurred numerous evaluation benchmarks, yet most rely on simulated environments or open-source applications, leaving real-world closed-source applications largely unevaluated. The core difficulty is that closed-source applications do not expose internal states, making traditional automatic verification inapplicable. To bridge this gap, we introduce \textbf{AndroidDaily}, a large-scale benchmark comprising 350 realistic daily-use tasks across 94 high-frequency Android applications spanning transportation, shopping, local services, entertainment, content creation, social media, and everyday utilities. To enable automatic and verifiable assessment in these opaque environments, we propose Guideline-grounded Reviewer for Automatic Diagnostic Evaluation (\textbf{GRADE}), a process-aware evaluator built on a three-tiered system of observable external guidelines: \emph{operational obligations}, \emph{output quality}, and \emph{negative constraints}. GRADE tracks the agent's visual trajectory against these criteria and produces step-level diagnostic judgments, turning long-horizon, open-ended mobile interactions into verifiable evaluation without relying on hidden internal states. Experiments show that GRADE achieves 87.37\% agreement with human evaluators. The strongest model reaches a 62.0\% success rate on AndroidDaily, highlighting a substantial gap between current reasoning capabilities and practical execution in realistic mobile workflows.

%% file: body/1-intro.tex
\begin{figure}[t]
  \centering
  \includegraphics[width=\columnwidth]{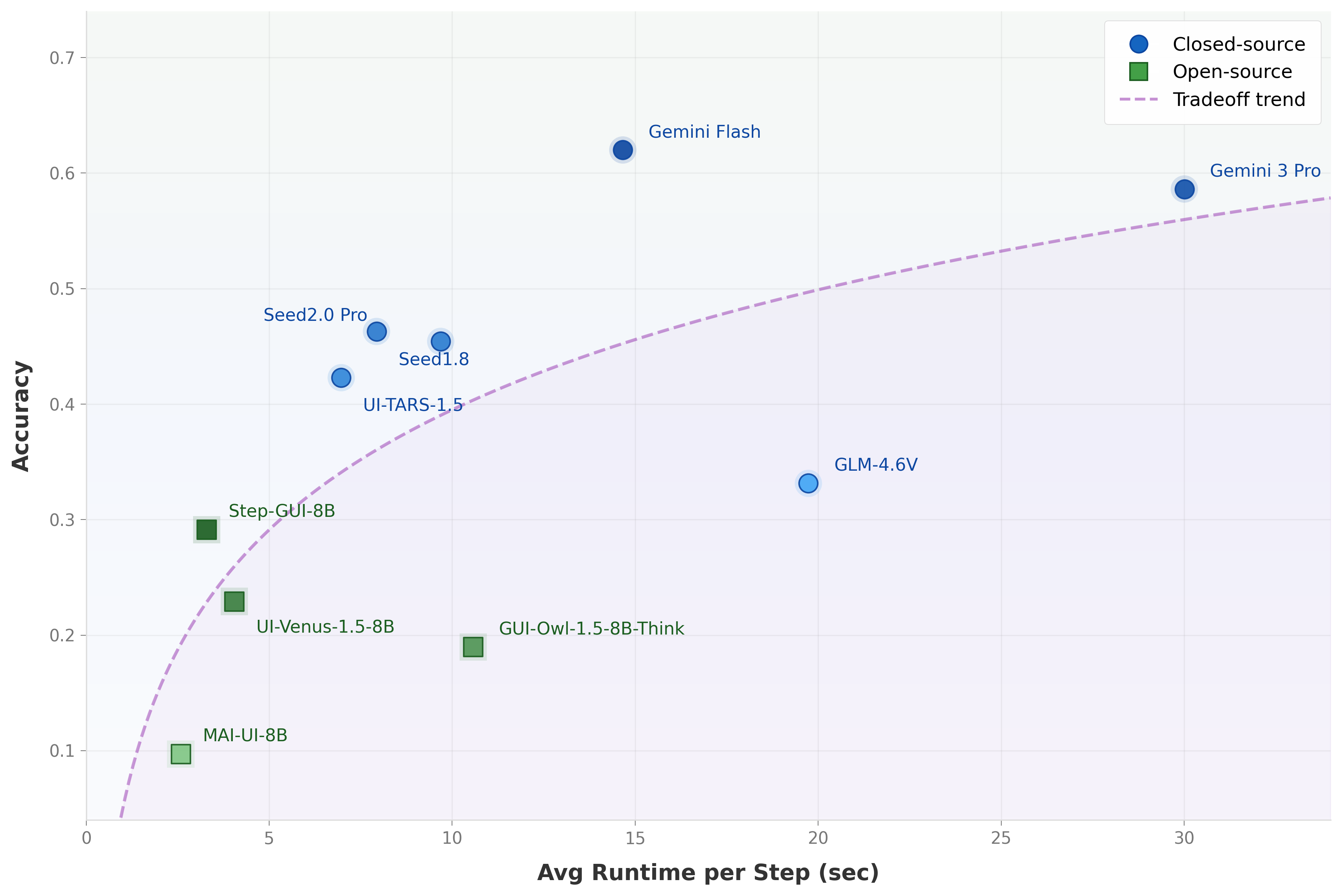}
  \caption{Relationship between per-step inference latency and task success rate on AndroidDaily. Closed-source models generally trade higher latency for better performance, while open-source models cluster at low latency but lower accuracy. Gemini 3 Flash achieves the best accuracy--latency tradeoff.}
  \label{fig:acc-latency}
\end{figure}

\section{Introduction}
\label{sec:intro}

GUI agents, intelligent systems capable of perceiving, understanding, and autonomously operating graphical user interfaces, hold the promise of transforming how users interact with mobile devices, from automating repetitive tasks to enabling hands-free accessibility~\cite{hong2024cogagent, qin2025ui, wang2024mobile}. As these systems move closer to real-world deployment, rigorous and systematic evaluation becomes essential for understanding their capabilities and guiding further improvement. Early efforts primarily target static settings, assessing an agent's ability to perceive and localize interface elements through tasks such as widget captioning, screen summarization, and element grounding~\cite{li2025screenspot,kapoor2024omniact}. More recently, a line of work has revealed a fundamental gap between static grounding proficiency and dynamic execution capability~\cite{li2026gui}, motivating the development of online benchmarks that evaluate agents through closed-loop interaction with live environments~\cite{rawles2024androidworld, xie2024osworld}. Despite this rapid progress, existing online benchmarks predominantly center on productivity or utility applications (Table~\ref{tab:benchmark_comparison}). A critical gap remains: the disconnect between evaluated tasks and actual daily usage patterns. Automated evaluation on real-world closed-source applications (the apps that dominate everyday mobile use) remains largely unexplored. Unlike AndroidWorld~\cite{rawles2024androidworld} and OSWorld~\cite{xie2024osworld}, which evaluate agents in open-source or sandboxed environments where success can be verified by injecting probes into application state, the commercial applications that dominate daily mobile use expose no such interface. Existing approaches therefore either restrict themselves to a handful of utility apps with privileged access, or fall back to coarse end-state matching that misses the long-horizon, multi-constraint behaviors characteristic of realistic usage.

\begin{figure*}[t]
  \centering
  \includegraphics[width=\textwidth]{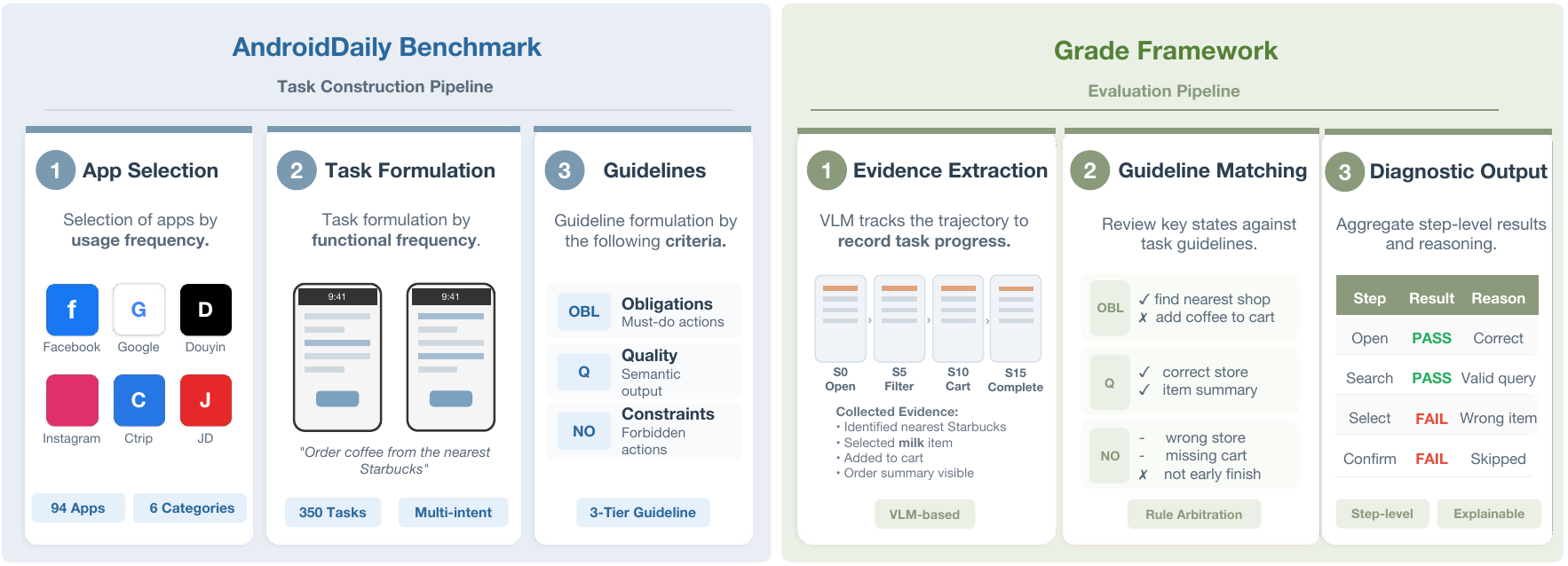}
  \caption{\textbf{Overview of AndroidDaily and GRADE.} 
  Left: the Task Construction Pipeline builds AndroidDaily with 350 realistic tasks across 94 high-frequency apps, each annotated with three-tier guidelines.
  Right: the GRADE Evaluation Pipeline evaluates agent trajectories against task guidelines and produces step-level diagnostic judgments.
  }
  \label{fig:overview}
\end{figure*}

To bridge this gap, we introduce AndroidDaily, a large-scale end-to-end mobile GUI agent benchmark grounded in empirical mobile usage patterns (Figure~\ref{fig:overview}). Based on usage frequency data and download statistics, we curate 350 realistic tasks spanning 94 high-frequency closed-source Android applications across key daily-life scenarios, including transportation, food delivery, shopping, social media, short-video consumption, entertainment, and local services. Unlike utility-focused benchmarks, our tasks involve real-world consequences such as financial transactions and service bookings, multi-step decision-making with competing constraints, and cross-application workflows that require coordinating information across multiple apps. These characteristics make AndroidDaily substantially more representative of the scenarios where agent deployment has immediate practical impact.

\input{tables/relatedbench}

Evaluating open-ended tasks on real-world closed-source applications introduces a fundamental challenge: internal application states (backend databases, complete UI trees, API responses) are entirely inaccessible, rendering code-based assertions of the kind employed by AndroidWorld and OSWorld inapplicable. Meanwhile, the tasks themselves admit inherent flexibility: a single goal may have multiple valid execution paths, different acceptable operation orderings, and varying legitimate stopping points. Together, these factors make deterministic automatic evaluation either infeasible or unreliable. To address this, we propose GRADE (Guideline-grounded Reviewer for Automatic Diagnostic Evaluation), a novel evaluation framework that decouples task verification from internal state access. For each task, we define a three-tiered system of external guidelines: (1)~\emph{operational obligations} that specify the necessary operations and constraints required for task completion, (2)~\emph{output quality} criteria that evaluate the quality of agent-generated content such as summaries, comparisons, or creative text, and (3)~\emph{negative constraints} that define actions strictly to be avoided, establishing safety and stopping boundaries. Acting as an LLM-based process-aware evaluator, GRADE tracks the agent's visual trajectory, matches execution steps against these guidelines, and produces a step-level diagnostic trace rather than a single binary pass/fail verdict, making it possible to pinpoint where and why an agent deviates from the task objective. 

Applying AndroidDaily with GRADE to a range of state-of-the-art GUI agent models reveals significant limitations in current capabilities (Figure~\ref{fig:acc-latency}). GRADE achieves 87.37\% agreement with human judgments, validating its reliability as an automatic evaluator for closed-source settings. The strongest current model reaches only 62.0\% overall success, with all models degrading sharply on multi-constraint and cross-app tasks. Our failure analysis identifies three dominant bottlenecks: \emph{inference latency} causing UI misalignment, \emph{memory-induced action loops}, and \emph{protocol-induced capability degradation}. These findings suggest that the key challenges for next-generation GUI agents lie not only in improving base capabilities, but in simultaneously achieving fast execution, robust history management, and reliable capability preservation under agentic workflows.

Our contributions are as follows:
\begin{itemize}
  \item We introduce \textbf{AndroidDaily}, a large-scale end-to-end mobile GUI agent benchmark grounded in real daily usage patterns. It comprises 350 tasks spanning 94 high-frequency closed-source Android applications across diverse everyday scenarios, emphasizing cross-app workflows, multi-constraint decision-making, and feedback-driven interaction.
  \item We propose \textbf{GRADE}, an automated process-aware evaluation framework tailored to closed-source environments. By leveraging a three-tiered system of observable guidelines (operational obligations, output quality, and negative constraints), GRADE enables verifiable and step-level diagnostic assessment without requiring access to internal application states.
  \item We conduct systematic experiments on AndroidDaily using state-of-the-art GUI agent models. Our results reveal substantial capability gaps under realistic daily-use settings and identify three dominant failure modes: latency-induced misalignment, memory-induced action loops, and protocol-induced capability degradation, establishing a baseline and roadmap for future research.
\end{itemize}

%% file: tables/relatedbench.tex

\begin{table}[t]
    \centering
    \small
    \setlength{\tabcolsep}{2pt}
    \resizebox{\columnwidth}{!}{%
        \begin{tabular}{lcccccc}
        \toprule
        \textbf{Benchmark} & \textbf{Language} & \textbf{Platform} & \textbf{Tasks} & \textbf{R} & \textbf{On} & \textbf{S} \\
        \midrule
        Mind2Web \cite{deng2023mind2web} & EN & Web & 2350 & \cmark & \xmark & \xmark \\
        ScreenSpot \cite{cheng2024seeclick} & EN & Mobile, Desktop, Web & 1272 & \cmark & \xmark & \xmark \\
        ScreenSpot Pro \cite{li2025screenspot} & EN & Desktop & 1581 & \cmark & \xmark & \xmark \\
        ScreenQA \cite{hsiao2025screenqa} & EN & Mobile & 85984 & \cmark & \xmark & \xmark \\
        MMBench-GUI \cite{wang2025mmbench} & EN & Mobile, Desktop, Web & 8123 & \cmark & \xmark & \xmark \\
        SPA-BENCH \cite{chen2024spa} & EN+CN & Mobile & 340 & \xmark & \cmark & \xmark \\
        AndroidWorld \cite{rawles2024androidworld} & EN & Mobile & 116 & \xmark & \cmark & \xmark \\
        OSWorld \cite{abhyankar2025osworld} & EN & Desktop & 369 & \cmark & \cmark & \xmark \\
        GUI-CEval \cite{li2026gui} & CN & Mobile & 8222 & \cmark & \cmark & \xmark \\
        \midrule
        \textbf{AndroidDaily} & {EN+CN} & {Mobile} & 350 & \cmark & \cmark & \cmark \\
        \bottomrule
        \end{tabular}
    }
    \caption{Comparison of existing GUI agent benchmarks with our AndroidDaily. R, On, S denote Real-word, Online, and Step-level
verification.}
    \label{tab:benchmark_comparison}
\end{table}

%% file: body/2-related-work.tex
\section{Related Work}
\label{sec:related}

\subsection{GUI Agents}

Recent advances in Multimodal Large Language Models (MLLMs) have driven rapid progress in autonomous GUI agents. Current approaches broadly fall into two paradigms. End-to-end models directly map visual observations and natural-language instructions to executable actions through a single unified model. Early explorations such as CogAgent~\cite{hong2024cogagent}, SeeClick~\cite{cheng2024seeclick}, and Auto-UI~\cite{zhan2023autoui}, together with related multimodal models such as Fuyu~\cite{fuyu-8b}, demonstrated the viability of this approach. More recent work, including UI-TARS~\cite{qin2025ui}, Step-GUI~\cite{yan2025step, yan2025gui}, Qwen3-VL~\cite{bai2025qwen3}, and Claude Opus 4.6~\cite{anthropic2026opus46}, has further advanced performance. Models such as Ferret-UI~\cite{you2024ferret} and Ferret-UI Lite~\cite{yang2025ferret} have improved mobile UI understanding and lightweight deployment, while open foundation action models such as OS-Atlas~\cite{wu2024atlas}, together with generalized automation models like EvoCUA~\cite{xue2026evocua} and GUI-Owl, introduced in Mobile-Agent-v3~\cite{ye2025mobile}, further broaden the applicability of this paradigm. In parallel, modular frameworks wrap off-the-shelf MLLMs in tool-use or multi-agent pipelines. On mobile, AppAgent~\cite{zhang2025appagent}, AutoDroid~\cite{wen2024autodroid}, DroidBot~\cite{li2017droidbot}, and the Mobile-Agent series~\cite{wang2024mobile,ye2025mobile} have advanced API-less interaction and exploration mechanisms, while desktop and web frameworks such as OS-Copilot~\cite{wu2024copilot}, UFO~\cite{zhang2025ufo}, Agent S~\cite{agashe2024agent}, Cradle~\cite{tan2024cradle}, WebVoyager~\cite{he2024webvoyager}, and CoAct~\cite{song2025coact} have demonstrated increasingly capable cross-application workflows, often aided by pixel-level parsing tools like OmniParser~\cite{wan2024omniparser}. Despite this progress in both paradigms, evaluating these agents on real-world closed-source commercial applications remains fundamentally difficult, as such apps provide no access to privileged internal APIs or underlying backend states.

\subsection{Evaluation of GUI Agents}

\noindent\textbf{Static evaluation.} Early work assessed GUI agents on static perception tasks: identifying interface elements, inferring semantics from single screenshots, or predicting isolated actions. Foundational datasets like Rico~\cite{deka2017rico} enabled tasks such as Widget Captioning~\cite{li2020widget} and screen summarization via Screen2Words~\cite{wang2021screen2words}. For action-oriented grounding, ScreenSpot~\cite{cheng2024seeclick} established a cross-platform benchmark for UI element localization, later extended by ScreenSpot-Pro~\cite{li2025screenspot} to complex professional software, while OS-Atlas~\cite{wu2024atlas} revised the original ScreenSpot annotations and released the corrected benchmark as ScreenSpot-V2. OmniACT~\cite{kapoor2024omniact} broadened static evaluation to desktop and web action generation. Large-scale trajectory datasets including AITW~\cite{rawles2023androidinthewild} and Mind2Web~\cite{deng2023mind2web} further enriched this line of research. While these benchmarks advanced basic visual perception and grounding, they only weakly capture real-world deployment challenges such as multi-step coordination and error recovery.

\begin{figure*}[t] \centering \includegraphics[width=\textwidth]{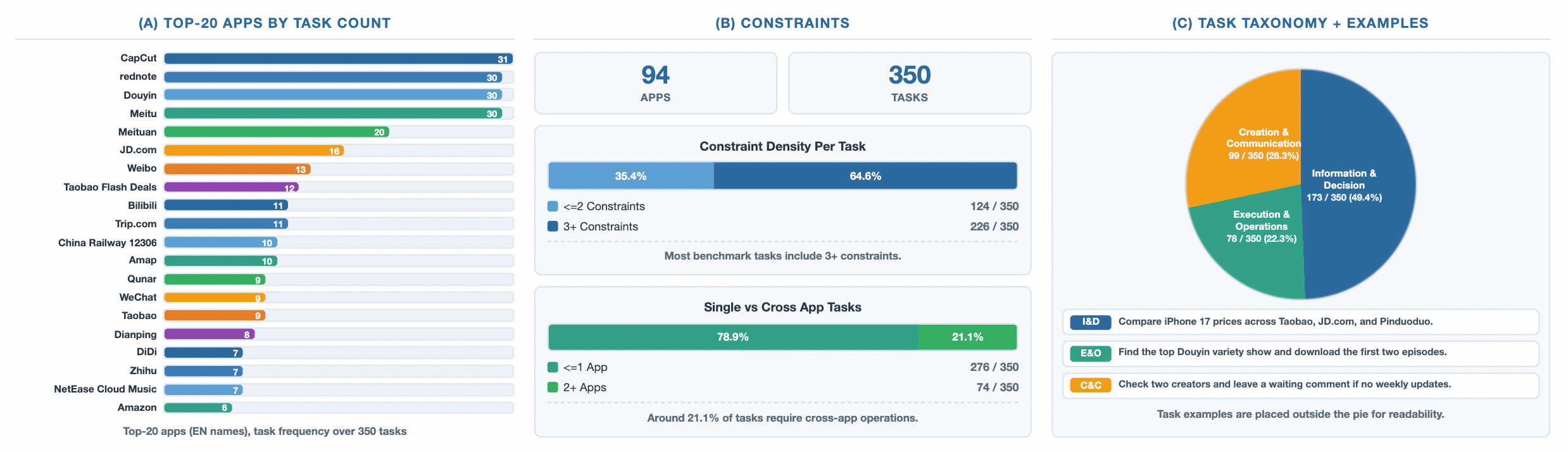} 
\caption{Benchmark statistics of AndroidDaily. (a) Top-20 apps by task frequency (English app names) over 350 tasks, led by CapCut (31), rednote (30), Douyin (30), and Meitu (30). (b) Benchmark scale and constraint density: 94 apps, 350 tasks, with <=2 constraints (124, 35.4\%) vs 3+ constraints (226, 64.6\%). (c) Task taxonomy pie with concise examples: Information \& Decision (173, 49.4\%), Execution \& Operations (78, 22.3\%), and Creation \& Communication (99, 28.3\%).} 
\label{fig:benchmark_stats} 
\end{figure*}

\noindent\textbf{Dynamic evaluation.} To address these limitations, the field shifted to online, closed-loop evaluation. Building on earlier synthetic environments such as MiniWoB++~\cite{liu2018reinforcement}, later work introduced more realistic execution-based environments for web and desktop interaction, including WebArena~\cite{zhou2023webarena} and OSWorld~\cite{xie2024osworld}, followed by VisualWebArena~\cite{koh2024visualwebarena}, BrowserGym~\cite{chezelles2025browsergym}, Windows Agent Arena~\cite{bonatti2024windows}, AssistGUI~\cite{gao2023assistgui}, CRAB~\cite{xu2024crab}, and MMBench-GUI~\cite{wang2025mmbench}. On mobile, AndroidWorld~\cite{rawles2024androidworld} offered dynamic evaluation with parameterized tasks, and more recent benchmarks such as MobileWorld~\cite{kong2025mobileworld} and GUI-CEval~\cite{li2026gui} expanded task difficulty, cross-app interaction, and app coverage. Recent efforts like OSWorld-Human~\cite{abhyankar2025osworld} further introduced human-efficiency baselines. Many of these benchmarks verify task completion through code-based assertions that inspect benchmark-maintained state, API responses, or system variables---an approach that is highly effective in open-source or simulated settings but hard to scale robustly to closed-source commercial apps, where backend states are hidden and evaluators must often rely on fragile UI-tree heuristics or superficial screen matching.

\noindent\textbf{LLM-based evaluation.} Beyond GUI-specific benchmarks, the broader NLP community has increasingly adopted LLM-based evaluation as a scalable alternative to human annotation. Pioneering work such as MT-Bench and Chatbot Arena~\cite{zheng2023judging} demonstrated that strong LLMs can serve as reliable pairwise judges for open-ended text generation. Subsequent efforts including JudgeLM~\cite{zhu2023judgelm}, which fine-tuned dedicated judge models to mitigate positional and knowledge biases, and CritiqueLLM~\cite{ke2024critiquellm}, which trained models to produce informative diagnostic critiques beyond scalar scores, have advanced the reliability and interpretability of automated evaluation. Comprehensive surveys~\cite{gu2024survey,li2024llms} have further systematized this rapidly growing paradigm. In a related direction, process reward models~\cite{lightman2023let} have demonstrated that step-level supervision can yield more faithful credit assignment than outcome-only scoring in mathematical reasoning. Recent work has extended this idea to agent settings, with AgentPRM~\cite{xi2025agentprm} introducing promise--progress decomposition for sequential decision-making, and ToolPRMBench~\cite{li2026toolprmbench} providing a dedicated benchmark for evaluating process-level rewards in tool-use scenarios. However, these ideas remain much less developed in GUI-agent evaluation than in text-only reasoning or general tool-use settings, especially for long-horizon trajectories on visually complex, closed-source commercial interfaces. Existing GUI benchmarks either rely on code-based assertions tied to privileged state access, or fall back to coarse binary judgments. GRADE bridges this gap by combining the LLM-as-a-Judge paradigm with process-aware trajectory tracking and externally defined three-tiered guidelines, enabling verifiable and diagnosable evaluation on closed-source commercial applications without requiring internal state access.

%% file: body/3-bench0329.tex
\section{Method}
\label{sec:method}

This section presents the design of AndroidDaily in detail. We begin with an overview of the benchmark scope and task construction principles (\S\ref{sec:benchmark}), describe the guideline-centric task specification that defines externally checkable completion criteria (\S\ref{sec:guideline}), and finally introduce GRADE, the process-aware evaluation framework that operationalizes these guidelines into automatic diagnostic assessment (\S\ref{sec:grade}).

\subsection{AndroidDaily Benchmark}
\label{sec:benchmark}

The core design principle of AndroidDaily is to closely mirror real-world mobile usage dynamics. We pursue this goal through two strategies: empirical application selection and diverse task formulation.

\noindent\textbf{Application Selection.} Rather than sampling from broad catalogs or relying on open-source alternatives, we ground our selection in empirical usage data to retain high-frequency closed-source applications that cover major daily-use categories. Based on app store download rankings and publicly available active-user statistics, we curate a set of 94 closed-source applications. These apps span the categories users most routinely engage with, including transportation, food delivery, e-commerce, social media, short-video platforms, entertainment, and local lifestyle services.


\noindent\textbf{Task Formulation.} For each app, we design tasks that reflect its core functions, common usage patterns, and recurring user intents. Instructions are formulated in natural language with varying levels of specificity to mimic real user behavior, ranging from \emph{precise directives} (e.g., ``order a luxury car from Location A to Location B'') to \emph{underspecified intents} (e.g., ``play something funny''), which require the agent to infer user preferences and navigate ambiguity.

\noindent\textbf{Task Allocation and Scale.} Dynamic GUI benchmarks typically contain a limited number of tasks due to the high cost of real-device execution (for instance, OSWorld includes 369 tasks and AndroidWorld 116 tasks). To achieve sufficient breadth while keeping total evaluation time manageable, we adopt a frequency-weighted allocation strategy: apps with higher daily usage frequency are assigned more tasks covering a wider range of functionalities, while lower-frequency apps are tested only on their most essential use cases. This ensures that evaluation effort is concentrated where it matters most. Figure~\ref{fig:benchmark_stats} illustrates the task distribution across the top-20 apps in AndroidDaily. The resulting benchmark comprises 350 tasks across 94 applications. Tasks are grouped into three capability-oriented categories:
\begin{itemize}[leftmargin=*]
  \item \emph{Information \& Decision}: Retrieving, comparing, and synthesizing information to support user decisions, ranging from short-horizon lookups to cross-app exploration.
  \item \emph{Creation \& Communication}: Generating user-visible content or engaging in semantic interactions such as composing posts, editing multimedia, commenting, and messaging.
  \item \emph{Execution \& Operations}: Completing concrete actions such as purchasing, booking, navigating, or adjusting settings, with an emphasis on step-by-step reliability and strict constraint satisfaction.
\end{itemize}

Together, these categories ensure that AndroidDaily is comprehensive in the range of capabilities it exercises: visual perception, intent understanding, long-horizon planning, cross-app coordination, content generation, and precise manipulation.

\subsection{Guideline-Centric Task Specification}
\label{sec:guideline}

\begin{figure*}[t]
  \centering
  \includegraphics[width=\textwidth]{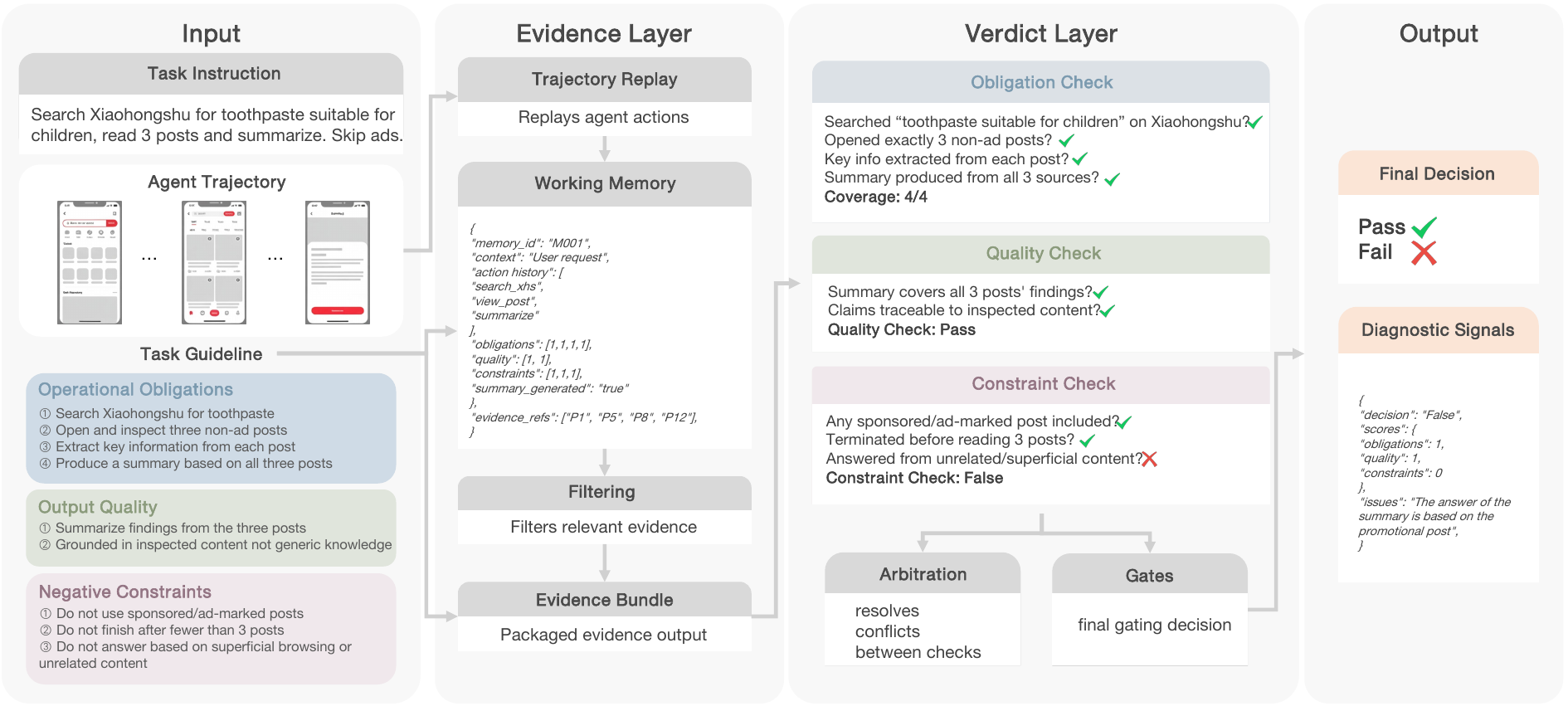}
  \caption{The GRADE evaluation pipeline. The Evidence Layer replays the agent trajectory to assemble structured evidence. The Verdict Layer checks obligations, quality, and boundaries to produce a pass/fail decision with diagnostic signals.}
  \label{fig:grade_ppl}
\end{figure*}

While AndroidDaily reflects realistic mobile usage, it also brings a practical challenge: most task instructions are open-ended and underspecified. Instead of providing explicit step-by-step commands, users may ask for things like ``\textit{Any recently trending TV shows? I want something to binge-watch.}'' or ``\textit{Help me find a place to take the kids on the weekend.}'', leaving the agent to infer missing constraints before acting. Evaluating task completion from the instruction alone is therefore inherently ambiguous. To address this, we pair each task with an external guideline that converts the open-ended instruction into explicit, observable completion conditions through three complementary tiers.

\noindent\textbf{Operational Obligations.} This tier specifies the core operations and intermediate states that must be satisfied for a task to be considered complete, regardless of the specific execution path taken. It encompasses target objects, key constraints, and necessary operations, providing the primary reference for judging basic task fulfillment.

\noindent\textbf{Output Quality.} This tier defines criteria for evaluating the quality of the agent's final state or generated content. Many daily-use tasks require the agent to produce textual outputs (e.g., summaries, recommendations) or specific visual outcomes. These criteria ensure that evaluation captures not just whether the agent performed the right actions, but whether its deliverables meet the task's semantic and informational goals.

\noindent\textbf{Negative Constraints.} This tier defines behaviors that must strictly be avoided, including boundary violations, invalid completion patterns, or safety hazards. A trajectory may appear successful on the surface yet fail if it violates a key constraint (for instance, executing a final payment that the user only asked to preview). Negative constraints prevent the evaluation from being misled by superficially plausible end states.

Together, these three tiers act as a standardized evaluation contract. By translating open-ended user intents into explicit, observable constraints, this specification effectively eliminates subjective ambiguity, ensuring high consistency and reliability in the subsequent automated evaluation.

\subsection{GRADE Evaluator}
\label{sec:grade}

Given the guideline, the remaining question is how to apply it to an agent's execution trajectory without access to internal application states. In realistic closed-source tasks, the terminal screen alone is insufficient: many decisive cues appear only briefly in intermediate steps and may later be overwritten. GRADE addresses this by operating as a \textbf{Vision-Language Model (VLM)-based} evaluator that takes the full multimodal trajectory (a sequence of screenshots and available accessibility metadata) alongside the external guideline (Figure~\ref{fig:grade_ppl}). It organizes the evaluation into two layers:

\noindent\textbf{Evidence Layer.} This layer tracks, filters, and compresses the visual trajectory into structured evidence. Concretely, GRADE maintains a working memory throughout the trajectory replay, visually extracting and recording task-relevant signals: action history, confirmed facts, target bindings, comparison states, and blocker signals. Uninformative steps are filtered out, yielding a structured \emph{evidence bundle}. This design ensures that transient but critical visual cues (such as a price comparison visible only on an intermediate screen) are retained for judgment.

\noindent\textbf{Verdict Layer.} This layer performs a structured review of the evidence bundle against the three-tiered guideline. \emph{Operational obligations} are checked for coverage; \emph{output quality} is checked for semantic adequacy; and \emph{negative constraints} are checked for boundary violations. Beyond item-level review, the verdict layer handles conflicting evidence through explicit arbitration rules (for instance, an unresolved blocker can enforce a failure decision even if other obligations appear satisfied).

Algorithm~\ref{alg:grade-2layer} specifies the complete two-layer pipeline. The evidence layer walks through the trajectory step by step, updating a working memory $M$ and accumulating step-level evidence $P$, then compiles a structured evidence bundle $B$. The verdict layer checks $B$ against each of the three guideline tiers and an arbiter combines the slot verdicts into the final decision $y$ together with diagnostic signals $D$. This design makes GRADE's reliability depend on two distinct capabilities: fine-grained visual recognition at the evidence layer, and structured reasoning over aggregated evidence at the verdict layer.

\begin{algorithm}[t]
\caption{GRADE: Two-Layer Evaluation Procedure}
\label{alg:grade-2layer}
\small
\begin{algorithmic}[1]
\Require Task instruction $x$; Guideline $G = \{g^{obl}, g^{qual}, g^{neg}\}$; Trajectory $\tau = \{(o_1,a_1),\dots,(o_T,a_T)\}$
\Ensure Final verdict $y \in \{\text{True},\text{False}\}$; Diagnostic signals $D$
\State $M \gets \emptyset$;\; $P \gets \emptyset$
\Statex \textbf{Evidence Layer}
\For{each $(o_t, a_t) \in \tau$}
  \State $p_t \gets \textsc{TraceEvidence}(x, G, o_t, a_t)$
  \State $M \gets \textsc{UpdateMemory}(M, p_t)$
  \State $P \gets P \cup \{p_t\}$
\EndFor
\State $B \gets \textsc{BuildEvidenceBundle}(x, G, \tau, M, P)$
\Statex \textbf{Verdict Layer}
\State $v^{obl} \gets \textsc{CheckObligations}(g^{obl}, B)$
\State $v^{qual} \gets \textsc{CheckOutputQuality}(g^{qual}, B)$
\State $v^{neg} \gets \textsc{CheckNegativeConstraints}(g^{neg}, B)$
\State $y \gets \textsc{Arbitrate}(v^{obl}, v^{qual}, v^{neg}, B)$
\State $D \gets \textsc{BuildDiagnostics}(v^{obl}, v^{qual}, v^{neg}, B)$
\State \Return $(y, D)$
\end{algorithmic}
\end{algorithm}

GRADE ultimately outputs a final completion decision alongside a set of diagnostic signals (e.g., obligation coverage, boundary violations, and failure-reason tags). This structured output makes decisions more stable than one-shot LLM judgments and supports downstream error analysis.

\vspace{1ex}
\noindent\textbf{An Open Evaluation Protocol.} Taken together, the task set, the guideline specification, and the GRADE evaluator form a modular and extensible paradigm. Because GRADE is task-agnostic and operates solely on the instruction--guideline--trajectory triplet, these components are not tightly coupled. Users can define new tasks on arbitrary apps, write custom guidelines, and run GRADE without any framework modification. AndroidDaily is therefore not just a fixed dataset, but an open evaluation protocol: flexible in task scope, customizable in evaluation standards, and applicable to any realistic mobile scenario.


%% file: body/4-experimental-setup.tex
\input{tables/main-results}

\section{Experiments}
\label{sec:experiments}

We evaluate multiple state-of-the-art GUI agent models on AndroidDaily, validate the reliability of GRADE against human annotators, and analyze the distinctive failure modes that emerge from large-scale evaluation.

\input{tables/grade_acc}

\subsection{Experimental Setup}
\label{sec:setup}

\noindent\textbf{Models.} We evaluate twelve mobile GUI agent models organized by training paradigm: \emph{general-purpose VLMs} adapted for GUI interaction and \emph{GUI-specialized models} trained with dedicated agent pipelines. The general-purpose group includes Gemini 3 Flash and Gemini 3 Pro~\cite{gemini3pro2025}, Seed1.8~\cite{bytedance2026seed18} and Seed2.0 Pro~\cite{bytedance2026seed20}, and GLM-4.6V~\cite{glmvteam2025glm4v}. The GUI-specialized group includes UI-TARS-1.5~\cite{qin2025ui} (a closed-source system), Step-GUI-8B and Step-GUI-30A3~\cite{yan2025step}, UI-Venus-1.5-8B~\cite{gao2026ui}, GUI-Owl-1.5-8B-Think and GUI-Owl-1.5-32B-Think~\cite{xu2026mobile}, and MAI-UI-8B~\cite{zhou2025mai}. This taxonomy is orthogonal to the open-/closed-source distinction and instead highlights the role of training paradigm: whether strong GUI performance emerges from general capabilities or requires domain-specific supervision.

\noindent\textbf{Rollout protocol.} For each model, we adopt its officially released inference protocol when available; for models without a public protocol, we tune the configuration to achieve the best performance before evaluation. Following the implementation of StepGUI~\cite{yan2025step}, all agent actions are executed on real Android devices via ADB. Task-specific environment preconditions (e.g., logged-in accounts, pre-loaded content, or required app states) are prepared before each session. During execution, a human monitor is present to intervene only in cases of unexpected device states such as screen-off or system-level interruptions, without influencing the agent's decisions. Each session is capped at a maximum of 60 interaction steps and 40 minutes of wall-clock time, and evaluated under the pass@1 setting. The full execution process is recorded as a chain-of-action trajectory indexed by a unique session ID, which serves as the input to GRADE for subsequent evaluation.

\subsection{Main Benchmark Results}
\label{sec:main-results}

\input{tables/reliability_single}

We report the overall benchmark results of representative mobile GUI agents on AndroidDaily. Table~\ref{tab:ad300x1-main-results} summarizes success rates across the full benchmark together with per-step latency and three objective slicing dimensions: guideline constraint count, app scope, and task taxonomy. For proprietary models, latency reflects API round-trip time; for locally deployed models, it reflects BF16 inference on H20 GPUs.

\noindent\textbf{Overall results.} AndroidDaily clearly separates model tiers. The top-performing systems are all general-purpose frontier VLMs: Gemini 3 Flash achieves the highest overall success rate of 62.0\%, followed by Gemini 3 Pro (58.6\%), Seed1.8 (46.3\%), and Seed2.0 Pro (45.4\%). GUI-specialized models lag substantially behind: the strongest is UI-TARS-1.5 at 42.3\%---a large-scale closed-source system trained with dedicated GUI agent pipelines including reinforcement learning---yet it still falls short of every general-purpose frontier model. The remaining GUI-specialized models drop further to 29.7\% (Step-GUI-30A3) and below. A closer examination of trajectory logs reveals a key qualitative difference: frontier models exhibit markedly stronger \emph{error-path recovery}. In real-device execution, agents routinely encounter environmental interference---slow page loads, unexpected pop-ups, landing on unintended screens, or dynamic content shifts. Frontier models recover from these disruptions more reliably by re-planning or backtracking, whereas GUI-specialized models tend to continue executing on the derailed path or enter repetitive loops. This robustness to real-world noise, rather than GUI grounding accuracy alone, accounts for much of the performance gap.

\noindent\textbf{Latency and reasoning depth.} As visualized in Figure~\ref{fig:acc-latency}, latency interacts with success rate differently across the two model groups. Among general-purpose frontier models, longer per-step inference time generally correlates with higher success rates: Gemini 3 Flash (14.7s/step, 62.0\%) outperforms Seed1.8 (7.9s/step, 46.3\%) and Seed2.0 Pro (9.7s/step, 45.4\%). This suggests that the additional computation is spent on deeper reasoning---constraint tracking, cross-app coordination, and recovery planning---which directly benefits complex daily-use tasks. The correlation breaks down at the extreme: Gemini 3 Pro spends 30.0s per step yet scores lower than Flash (58.6\% vs.\ 62.0\%), because its high per-step latency accumulates on long-horizon tasks, frequently pushing sessions past the 40-minute time limit.

Among GUI-specialized models, however, no such latency--accuracy correlation is observed. UI-TARS-1.5, the strongest in this group, achieves 42.3\% at just 7.0s/step, yet GUI-Owl-32B is more than three times slower (25.6s/step) while scoring only 21.6\%. Within model families, scaling yields negligible gains: Step-GUI improves by just 0.6pp from 8B to 30A3, and GUI-Owl by 2.6pp from 8B to 32B. The additional parameters or computation do not translate into meaningfully better task completion. This contrast suggests that the reasoning patterns underlying frontier models' success---error recovery, constraint satisfaction under ambiguity, long-horizon re-planning---are not well captured by current GUI-specialized training pipelines. Closing this gap likely requires richer training data that emphasizes these capabilities, rather than parameter scaling alone.

\noindent\textbf{Slice-level analysis.} Even frontier models are far from solving AndroidDaily. Performance degrades consistently as task complexity increases. On the constraint dimension, every model drops from $\leq 2$ to $\geq 3$ constraints; for example, Gemini 3 Flash falls from 69.4\% to 58.0\%, and Seed1.8 from 58.9\% to 39.4\%. The app scope dimension reveals an even sharper decline: Gemini 3 Pro drops from 63.0\% to 41.9\% on multi-app tasks. Across the task taxonomy, Creation \& Communication tasks are generally harder than Execution \& Operations, particularly for GUI-specialized models (Step-GUI-8B: 23.2\% vs.\ 28.2\%), likely because they require generating contextually appropriate content beyond GUI manipulation. Overall, AndroidDaily not only separates models by aggregate success rate, but also systematically exposes weaknesses as tasks become more compositional and realistic---even for the strongest frontier systems.

\begin{table}[t]
  \centering
  \small
  \setlength{\tabcolsep}{4pt}
  \caption{\textbf{Backbone comparison for GRADE} on 879 manually reviewed sessions. Only the backbone model inside GRADE is changed across rows; the protocol, guidelines, and decision logic are held fixed.}
  \label{tab:evaluator-backbones}
  \begin{tabular}{lcccccc}
  \toprule
  Evaluator Backbone & $N$ & Acc.\,(\%) & TP & TN & FP & FN \\
  \midrule
  Gemini 3 Pro   & 879 & \textbf{87.37} & 188          & \textbf{580} & \textbf{86} & 25          \\
  Gemini 3 Flash & 879 & 80.55          & 193          & 515          & 151         & 20          \\
  Seed1.8        & 879 & 83.73          & 175          & 561          & 105         & 38          \\
  Seed2.0 Pro    & 879 & 84.19          & \textbf{195} & 545          & 121         & \textbf{18} \\
  GPT-4o         & 879 & 65.64          & 190          & 387          & 279         & 23          \\
  \bottomrule
  \end{tabular}
\end{table}

\begin{figure*}[t]
  \centering
  \includegraphics[width=\textwidth]{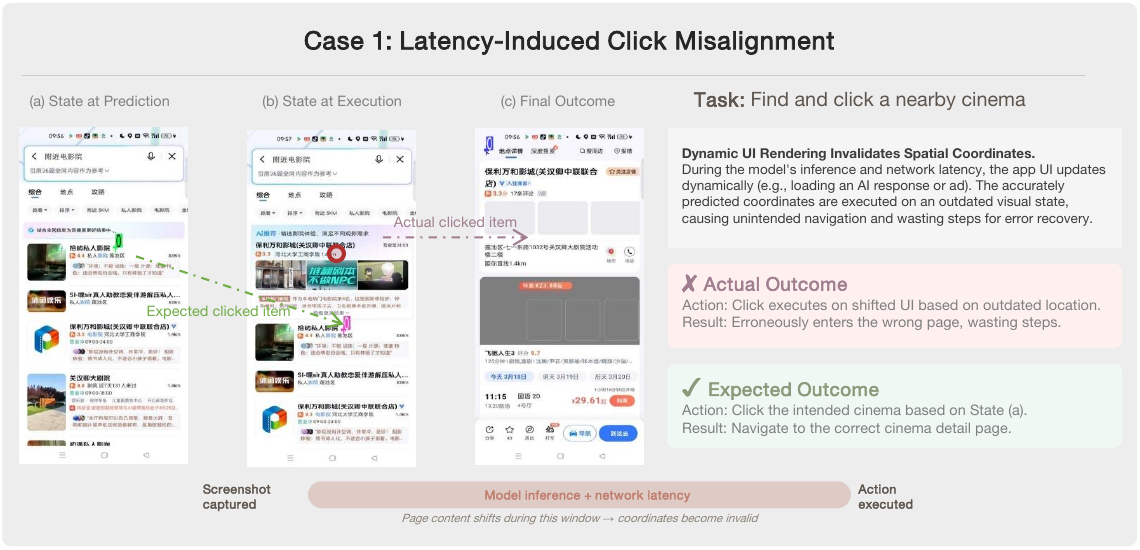}
  \caption{\textbf{Illustration of Latency-Induced Click Misalignment.}
 (a) At the prediction state, the model accurately plans spatial coordinates (green box) to click the target cinema. (b) During the execution phase, model inference and network latency allow the dynamic UI to update (e.g., inserting an AI response), shifting the intended target downward. The agent executes the click at the outdated coordinates, mistakenly hitting a different item (red circle). (c) The final outcome shows the agent navigating to an unintended page, which derails the task and wastes subsequent recovery steps.}
  \label{fig:case1}
\end{figure*}

\subsection{Evaluator Reliability}
\label{sec:reliability}

To assess GRADE's evaluation quality, we collected 879 sessions from the raw inference outputs across all baseline models. Each session was manually reviewed and calibrated through three successive rounds of human annotation to produce a reliable ground-truth label. This adjudicated label set serves as the reference for all subsequent reliability analyses.

\noindent\textbf{Layer ablation.} We first ablate GRADE's two-layer design. As shown in Table~\ref{tab:grade-acc}, the evidence layer alone already achieves 84.76\% accuracy (TP=185, TN=560, FP=106, FN=28). Adding the verdict layer primarily suppresses false positives from 106 to 86, while slightly improving true positives (188) and true negatives (580). The full two-layer pipeline reaches 87.37\% accuracy, confirming that both layers contribute meaningfully: the evidence layer provides a robust baseline by extracting structured trajectory evidence, while the verdict layer refines it by performing fine-grained constraint checking that catches trajectories that superficially appear successful but miss critical details.

\noindent\textbf{Backbone comparison.} To verify that GRADE's reliability stems from the protocol rather than the choice of a single backbone, we swap five VLMs into the same GRADE procedure (Table~\ref{tab:evaluator-backbones}). The backbone's visual perception capability has a decisive impact on verification accuracy. While true-positive counts remain stable across all backbones (175--195), false-positive counts vary dramatically---from 86 (Gemini 3 Pro) to 279 (GPT-4o). The vast majority of these false positives trace back to the same root cause: the backbone fails to detect a subtle error in the trajectory (e.g., a wrong item selected, a constraint silently unmet) and propagates this misperception into an incorrect positive verdict.

\noindent\textbf{Human agreement on hard cases.} This pattern is not unique to models. We selected 100 visually ambiguous trajectories and recruited 5 non-expert evaluators to independently judge their success. Table~\ref{tab:reliability_single} records pairwise agreement among all raters, including GRADE, the calibrated expert ground truth (GT), and the five annotators, under both with- and without-guideline settings. The results show that human evaluators struggle with the same root cause as model backbones: fine-grained visual details that determine task success are easy to overlook. Without guidelines, individual human--GT agreement ranges from just 49\% to 60\%; with guidelines it improves to 56--67\%, primarily by reducing false positives---annotators become less likely to accept superficially plausible trajectories that actually miss key constraints. Across both settings, GRADE maintains 72--73\% agreement with GT, consistently matching or exceeding every individual non-expert annotator. The difference is that models occasionally \emph{cannot perceive} critical visual cues, while humans \emph{do not attend to them carefully enough}---but the practical effect on evaluation reliability is similar. GRADE thus serves as a consistent and scalable automatic proxy for large-scale evaluation on AndroidDaily. 

\begin{figure*}[t]
  \centering
  \includegraphics[width=\textwidth]{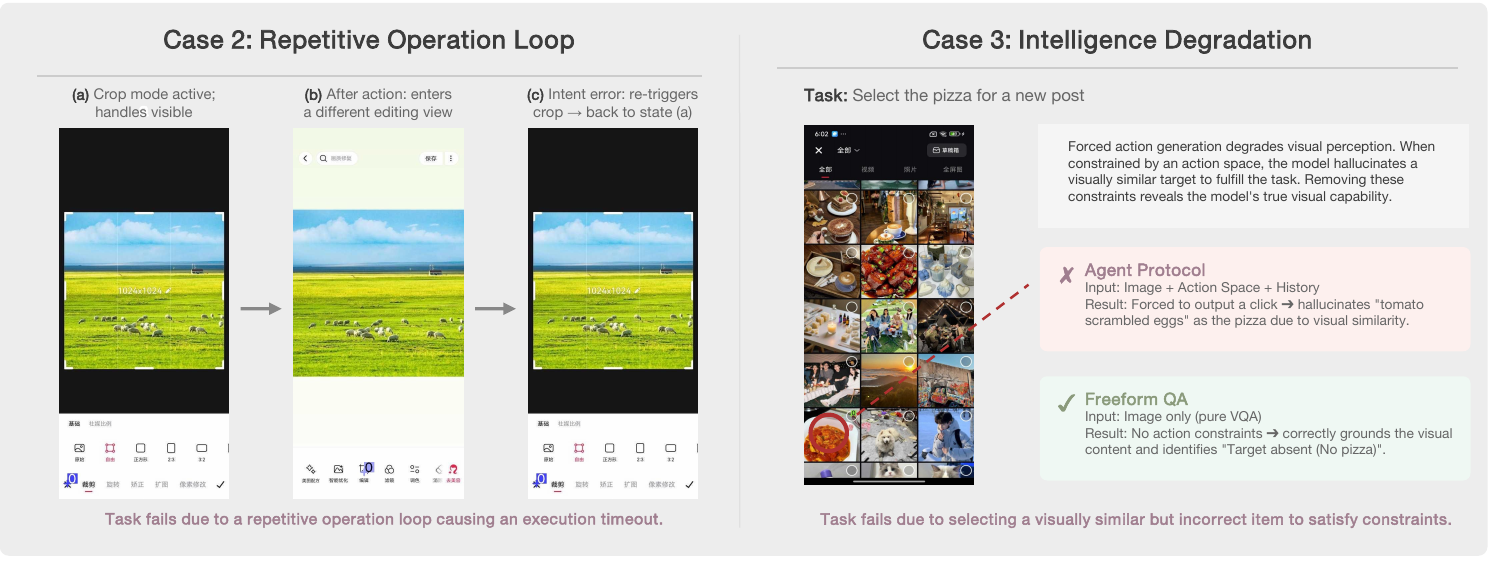}
  \caption{\textbf{Two more failure cases under realistic mobile evaluation.}  (Left) Repetitive Operation Loop: The agent becomes trapped in a cyclic execution path. After transitioning from the initial state (a) to a new view (b), an intent error causes it to revert to the same initial state (c), creating an infinite loop that results in an execution timeout. (Right) Intelligence Degradation: Augmenting the input with agent-specific context (action space definitions, interaction history, etc.) can probabilistically degrade the model's visual perception. When tasked to select ``pizza'' (absent from the screen), a pure VQA query correctly identifies its absence, whereas the same model under the agent protocol hallucinates a click on a visually similar item (``tomato scrambled eggs''), causing task failure.}
  \label{fig:case23}
\end{figure*}

\subsection{Failure Mode Analysis}
\label{sec:failure-analysis}

We summarize the most common failure patterns observed on AndroidDaily. To ground this analysis quantitatively, we manually categorized all failed sessions from the top three frontier and top three GUI-specialized models, identifying four dominant failure types.

\noindent\textbf{Frontier models: latency as the dominant bottleneck.} Despite stronger reasoning and recovery capabilities, frontier models are primarily limited by high inference latency, which manifests in two ways. First, \emph{timeout failures}: on long-horizon tasks ($\geq 3$ constraints or $\geq 3$ apps), accumulated per-step latency frequently pushes sessions past the 40-minute wall-clock limit, accounting for the sharp performance drop observed in Table~\ref{tab:ad300x1-main-results} for Gemini 3 Pro on these slices. Second, \emph{action misalignment with dynamic interfaces}: as shown in Figure~\ref{fig:case1}, the agent decides to click a target element, but during the delay between decision and execution, a transient UI element (e.g., an AI recommendation overlay, a pop-up, or a loading screen) appears and shifts the intended click target. The tap lands on the wrong element, forcing the agent to dismiss it and navigate back, wasting additional steps and compounding the latency problem.

\noindent\textbf{GUI-specialized models: decision quality as the primary weakness.} GUI-specialized models are faster but make substantially more decision errors, with two patterns standing out. First, \emph{memory-induced action loops}: as shown in Figure~\ref{fig:case23} (left), the agent loses track of its history and cycles between a few states indefinitely, exhausting the step budget without progress. Second, \emph{protocol-induced capability degradation}: in Figure~\ref{fig:case23} (right), the agent misidentifies scrambled eggs with tomatoes as pizza during GUI execution, yet correctly refuses to click ``pizza'' when the same image is presented in a free-form VQA setting. The perceptual capability exists but fails to activate under the structured action protocol, suggesting that agent-specific context (action space definitions, interaction history, structured output format) can probabilistically interfere with the underlying VLM's visual reasoning.

\noindent\textbf{Implications for future development.} These failure patterns point to three concrete research directions: (i)~\emph{latency-aware agent design}, including speculative action planning, asynchronous execution, or lightweight verification modules that can handle dynamic UI changes without full model re-inference; (ii)~\emph{robust long-horizon memory}, such as explicit state tracking or trajectory summarization mechanisms that prevent action loops; and (iii)~\emph{capability-preserving agentic protocols} that maintain the base VLM's perceptual accuracy when augmented with agent-specific context. The contrasting failure profiles of frontier and GUI-specialized models further suggest a fourth direction: (iv)~\emph{training data that captures recovery and re-planning patterns}, which frontier models appear to acquire from general pretraining but GUI-specialized pipelines currently lack. AndroidDaily, by requiring reasoning depth, fast response, and robust capability preservation simultaneously, provides a diagnostic testbed for progress along all four axes.

%% file: tables/main-results.tex
\begin{table*}[t]
\centering
\small
\setlength{\tabcolsep}{6pt}
\caption{\textbf{Main benchmark results on AndroidDaily.} Task success rates (\%) are reported by constraint count, app scope, and task taxonomy. Runtime is the average seconds per step over valid sessions. Models are grouped by training paradigm: general-purpose VLMs vs.\ models trained with dedicated GUI agent pipelines. Best results within each group are \textbf{bolded}.}
\label{tab:ad300x1-main-results}
\begin{tabular}{l cc cc cc ccc}
\toprule
& & & \multicolumn{2}{c}{\textit{Constraints}} & \multicolumn{2}{c}{\textit{App Scope}} & \multicolumn{3}{c}{\textit{Task Taxonomy}} \\
\cmidrule(lr){4-5} \cmidrule(lr){6-7} \cmidrule(lr){8-10}
Model & Avg Step Time (s) & Overall & $\leq$2 & 3+ & $\leq$1 App & 2+ Apps & Info.\&Dec. & Crea.\&Com. & Exec.\&Ops. \\
\midrule
\rowcolor{gray!15}
\multicolumn{10}{l}{\textit{General-purpose VLMs}} \\
\quad Gemini 3 Flash~\cite{gemini3flash2025}      & 14.7          & \textbf{62.0} & 69.4          & \textbf{58.0} & 62.7 & \textbf{59.5} & \textbf{63.6} & \textbf{61.6} & 59.0 \\
\quad Gemini 3 Pro~\cite{gemini3pro2025}      & 30.0          & 58.6          & \textbf{74.2} & 50.0          & \textbf{63.0}          & 41.9          & 60.7          & 50.5          & \textbf{64.1} \\
\quad Seed1.8~\cite{bytedance2026seed18}            & 7.9           & 46.3          & 58.9          & 39.4          & 48.9          & 36.5          & 47.4          & 46.5          & 43.6 \\
\quad Seed2.0 Pro~\cite{bytedance2026seed20}            & 9.7           & 45.4          & 55.6          & 39.8          & 46.0          & 43.2          & 46.8          & 44.4          & 43.6 \\
\quad GLM-4.6V~\cite{glmvteam2025glm4v}          & 19.7          & 33.1          & 43.5          & 27.4          & 35.9          & 23.0          & 36.4          & 23.2          & 38.5 \\
\midrule
\rowcolor{gray!15}
\multicolumn{10}{l}{\textit{GUI-specialized models}} \\
\quad UI-TARS-1.5~\cite{qin2025ui}       & 7.0  & \textbf{42.3}          & \textbf{54.0}          & \textbf{35.8}          & \textbf{47.1}          & \textbf{24.3}          & \textbf{41.6}          & \textbf{43.4}          & \textbf{42.3} \\
\quad Step-GUI-30A3~\cite{yan2025step}           & 2.0  & 29.7 & 41.9 & 23.0 & 33.0 & 17.6 & 31.8 & 23.2 & 33.3 \\
\quad Step-GUI-8B~\cite{yan2025step}           & 3.3  & 29.1 & 42.7 & 21.7 & 33.3 & 13.5 & 32.9 & 23.2 & 28.2 \\
\quad UI-Venus-1.5-8B~\cite{gao2026ui}          & 4.0           & 22.9          & 31.5          & 18.2          & 26.8          & 8.2           & 24.4          & 18.2          & 25.6 \\
\quad GUI-Owl-1.5-32B-Think~\cite{xu2026mobile}            & 25.6          & 21.6          & 33.6          & 15.1          & 25.5          & 7.4           & 29.7          & 9.3           & 20.3 \\
\quad GUI-Owl-1.5-8B-Think~\cite{xu2026mobile}            & 10.6          & 19.0          & 29.5          & 13.3          & 22.3          & 6.8           & 23.7          & 10.3          & 19.2 \\
\quad MAI-UI-8B~\cite{zhou2025mai}         & 2.6             & 9.7           & 15.3          & 6.6           & 12.0          & 1.4           & 8.7           & 13.1          & 7.7 \\
\bottomrule
\end{tabular}
\end{table*}

%% file: tables/grade_acc.tex
\begin{table}[t]
  \centering
  \small
  \caption{\textbf{Layer ablation of GRADE} on 879 manually reviewed sessions. Adding the verdict layer primarily reduces false positives.}
  \label{tab:grade-acc}
  \begin{tabular}{lcccccc}
  \toprule
  Configuration & $N$ & Acc.\,(\%) & TP & TN & FP & FN \\
  \midrule
  Evidence Layer only    & 879 & 84.76 & 185 & 560 & 106 & 28 \\
  \quad + Verdict Layer  & 879 & 87.37 & 188 & 580 &  86 & 25 \\
  \bottomrule
  \end{tabular}
\end{table}

%% file: tables/reliability_single.tex
\begin{table}[t]
\centering
\footnotesize
\caption{\textbf{Pairwise agreement (\%) on the 100-session challenging subset.} Each cell reports \textcolor{black}{with} / \textcolor{gray}{without} guideline. Only the lower triangle is shown.}
\label{tab:reliability_single}
\begin{tabular}{lccccccc}
\toprule
 & GT & GRADE & H1 & H2 & H3 & H4 & H5 \\
\midrule
GRADE   & 73\,/\,\textcolor{gray}{72} &  &  &  &  &  &  \\
H1      & 63\,/\,\textcolor{gray}{60} & 56\,/\,\textcolor{gray}{62} &  &  &  &  &  \\
H2      & 67\,/\,\textcolor{gray}{54} & 64\,/\,\textcolor{gray}{56} & 68\,/\,\textcolor{gray}{68} &  &  &  &  \\
H3      & 61\,/\,\textcolor{gray}{57} & 56\,/\,\textcolor{gray}{67} & 64\,/\,\textcolor{gray}{71} & 70\,/\,\textcolor{gray}{77} &  &  &  \\
H4      & 56\,/\,\textcolor{gray}{50} & 59\,/\,\textcolor{gray}{56} & 69\,/\,\textcolor{gray}{68} & 63\,/\,\textcolor{gray}{74} & 69\,/\,\textcolor{gray}{73} &  &  \\
H5      & 63\,/\,\textcolor{gray}{49} & 60\,/\,\textcolor{gray}{63} & 72\,/\,\textcolor{gray}{71} & 74\,/\,\textcolor{gray}{81} & 76\,/\,\textcolor{gray}{82} & 79\,/\,\textcolor{gray}{79} &  \\
\bottomrule
\end{tabular}
\end{table}

%% file: body/5-limitations.tex
\section{Limitations and Ethics}
\label{sec:limitations}

\noindent\textbf{Evaluation constraints.}
Following established practice in real-device GUI agent benchmarks~\cite{rawles2024androidworld,xie2024osworld}, we report \emph{pass@1} success rates without multi-seed variance, as each rollout requires up to 40 minutes of wall-clock time on physical Android hardware.
Multi-seed variance estimation on a representative subset is a valuable direction for future work.
In addition, the commercial applications underlying AndroidDaily evolve continuously through UI updates, A/B testing, and personalized content.
We mitigate environmental drift through controlled APK versioning and standardized account states (\S\ref{sec:setup}), but absolute success rates should be interpreted as a snapshot at evaluation time; relative comparisons across models evaluated in the same window remain meaningful.

\noindent\textbf{Ethics statement.}
All rollouts are executed on dedicated research-purpose accounts owned by the authors; no end-user accounts or third-party data are involved.
The \emph{negative constraints} tier of our guideline specification (\S\ref{sec:guideline}) is designed in part to prevent agents from completing irreversible real-world actions such as final payments or messages to unrelated parties, and all sessions are monitored by a human operator who can abort boundary-approaching runs.
Screenshots in the paper have been reviewed and personally identifiable information has been redacted.
We release task definitions, guidelines, and evaluation code rather than raw trajectory data, and encourage downstream users to comply with the applicable terms of service when reusing the protocol on new applications.
Extending coverage to professional, enterprise, and accessibility-specific workflows remains a natural direction for future work.

%% file: body/5-conclusion.tex
\section{Conclusion}
\label{sec:conclusion}

We presented AndroidDaily, a large-scale benchmark of 350 realistic daily-use tasks across 94 closed-source Android applications, together with GRADE, a process-aware evaluator that enables verifiable and diagnosable assessment without access to internal application states. GRADE achieves 87.37\% agreement with calibrated human labels. The strongest current model reaches only 62.0\% success, with all models degrading sharply on multi-constraint and cross-app tasks. Our analysis identifies three dominant failure modes: latency-induced misalignment, memory-induced loops, and protocol-induced capability degradation, pointing to the need for agents that combine reasoning depth, fast execution, and robust capability preservation under agentic protocols. AndroidDaily is designed as an open evaluation protocol and can be freely extended with custom tasks and guidelines.